# Internal Structure Attention Network for Fingerprint Presentation Attack Detection from Optical Coherence Tomography

Haohao Sun, Yilong Zhang, Peng Chen, *Member, IEEE* , Haixia Wang, *Member, IEEE*, and Ronghua Liang, *Senior Member, IEEE*

*Abstract*—As a non-invasive optical imaging technique, optical coherence tomography (OCT) has proven promising for automatic fingerprint recognition system (AFRS) applications. Diverse approaches have been proposed for OCT-based fingerprint presentation attack detection (PAD). However, considering the complexity and variety of PA samples, it is extremely challenging to increase the generalization ability with the limited PA dataset. To solve the challenge, this paper presents a novel supervised learning-based PAD method, denoted as ISAPAD, which applies prior knowledge to guide network training and enhance the generalization ability. The proposed dual-branch architecture can not only learns global features from the OCT image, but also concentrate on layered structure feature which comes from the internal structure attention module (ISAM). The simple yet effective ISAM enables the proposed network to obtain layered segmentation features belonging only to Bonafide from noisy OCT volume data directly. Combined with effective training strategies and PAD score generation rules, ISAPAD obtains optimal PAD performance in limited training data. Domain generalization experiments and visualization analysis validate the effectiveness of the proposed method for OCT PAD.

## I. Introduction

FINGERPRINTS have become the most widely used biometric features in current individual identification and authentication due to its safety, reliability and accuracy. A large number of automatic fingerprint recognition systems (AFRSs) are used for identity authentication and access control [1], [2]. However, presentation attacks (PAs, also referred as spoofing) are the most common and serious security issue for AFRS [3]. In particular, with the update of technology and processes, intruders can easily obtain and provide a variety of fake fingerprint samples to gain spurious access rights [4], [5]. With the widespread use of mobile devices and quick payment methods, the need to prevent PAs increasingly urgent due to the possibility of spoofing AFRS poses potential risks to users' privacy and financial security. However, the materials of PA are complex and diverse. It is difficult to build a complete database for PA detection (PAD) methods deployed in the public domain. Therefore, ensuring generalization capability (refers to the ability to accurately classify new, unseen data based on what it has learned from the training data) while being compatible with AFRS is a prerequisite for the generalization and adoption of one novel PAD technology.

Primary studies on PAD method can be roughly divided into two perspectives according to different implementation methods. The first perspective is based on pure software to extract PAD features, such as texture-based features [6], [7], physiological features [8], distortion information [9], and so on. The use of these features requires the combination of different classifiers. In addition, the method based on convolutional neural network (CNN) model has also been studied in depth [11]. Generally, the performance of software-based PAD methods is determined by the distinguishability of the PAD feature. In second perspective, additional detection techniques are incorporated into AFRSs to better complete the PAD. These methods include time-series detection of perspiration [12], sweat pores [13], pulse and oximetry [14], reaction to electric pulse [15], odor [16], multispectral imaging [17], combination of direct shooting and total reflection acquisition [18].

Although the existing hardware and software PAD schemes can solve PAs to a certain extent, it is difficult for these solutions to play a stable performance in field deployment due to the limited and unstable information obtained by the traditional acquisition equipment. Software-based PAD methods primarily use fingerprint images with only epidermal information. Uncontrollable hand conditions (dry, wet, wear, etc.) can directly affect its performance [19]. Besides when testing software methods on PAs of unknown material types, the performance drops rapidly [20]. Hardware-based PAD methods vary widely in hand conditions among different subjects, making it difficult in accurately modeling all live subjects [21]. In particular, ultra-thin PAs are likely to be ignored by additional detection to obtain vital information about intruders [21], [22].

This work was supported in part by National Natural Science Foundation of China under Grant 62276236, U1909203, 61976189. *(Corresponding author: Yilong Zhang.)*

H. Sun is with the College of Information and Engineering, Zhejiang University of Technology, Hangzhou 310023, China (e-mail: hhsun@zjut.edu.cn).

Y Zhang, P Chen, H Wang, Y Liu and R Liang are with the College of Computer Science & College of Software, Zhejiang University of Technology, Hangzhou 310023, China (e-mail: zhangyilong@zjut.edu.cn; chenpeng@zjut.edu.cn; hxwang@zjut.edu.cn; rhliang@zjut.edu.cn ).

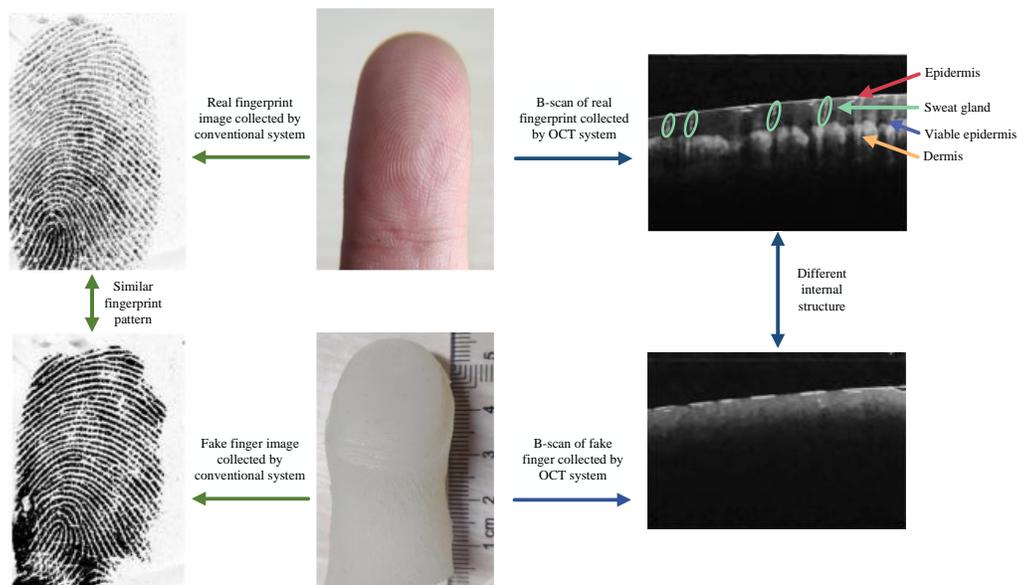

Fig. 1. Comparison between traditional fingerprint images and OCT-based internal structure data (B-scan). Real fingers and PA (produced by PDMS) with the same surface pattern can obtain similar fingerprint images on traditional fingerprint system. However, there are large differences in the internal structure data based on OCT. Real fingers have a layered internal structure, while PA has less internal structure. This difference is the basis for the OCT fingerprint system to implement PAD.

The inadequacy of traditional acquisition equipment prompted researchers to develop and use a new technology, optical coherence tomography (OCT) technology [23], [24], into fingerprint recognition systems. As an emerging optical technology, OCT has developed rapidly in recent years because of its real-time, 3D, high-sensitivity and label-free advantages [25], [26]. OCT fingerprint system can obtain finger subcutaneous tissue information [26], [27] and reconstruct internal finger images [28], [29]. Growing academic and research interest is attracted by OCT fingerprint systems as they offer more uniform fingerprint quality [30], fingerprint repair potential to traditional 2D fingerprint [31], and robust PAD capabilities [32].

Early studies reported that OCT fingerprint systems are compatible with traditional AFRSs. Aum et al. used the En-face method to reconstruct the internal fingerprint, in which the 48 of 50 samples were successfully matched with the traditional external fingerprint [33]. Sekulska-Nalewajko et al. found that OCT internal fingerprints have the same texture detail as external fingerprints based on visual evaluation [34]. Liu et al. verified that OCT fingerprints have better matching potential by fusing the fingerprints of various internal layers [24]. Sun et al. built a synchronous acquisition system that combines traditional fingerprint collection technology with OCT [23]. These studies on the correlation between OCT fingerprint and traditional fingerprint have proved that OCT fingerprint has broad application prospects.

Compared with traditional AFRSs, the configuration of the PAD method on the OCT fingerprint system is quite different. As shown in Fig. 1, traditional fingerprint acquisition systems need to implement PAD from similar pattern information. In contrast, the PAD of OCT fingerprint system can be realized by comparing the internal structure difference between real finger and PA. Internal structures in cross-section image (called B-scan) captured by OCT, such as sweat glands, viable epidermal junction and dermis, can provide foundations for PAD as unique features. In fact, while forging the same epidermal pattern, it is almost impossible for current PA technological capability to counterfeit of the corresponding internal sweat glands and dermis.

OCT-based fingerprint PAD methods can be divided into perspectives: feature-based methods and learning-based methods. In early study, OCT-based PAD was realized by manually screening image features. Bossen et al. detected PA by determining the existence of internal fingerprints in OCT volume data [35]. Liu et al. proposed that PA can be judged based on sweat glands using OCT [36]. While these manual methods are intuitive and effective, given the high labor costs, automated PAD is clearly a more adaptable approach to the system configuration. The automatic feature-based approach focuses on differences between bonafide and PA in OCT internal cross-section image. Sousedik et al. detected the boundary between layers and used the energies of the detected layers for classification [29]. Darlow et al. proposed a combination strategy using finger internal layered depth representation and autocorrelation analysis for PAD [37]. Liu et al. realized the successful detection of multiple types of spoofing attacks by peak feature (called depth-double-peak feature and sub-single-peak feature) [32]. These feature-based approaches both have complex processing flow and are time consuming. More importantly, it is difficult to build a unified model to characterize complex PAs.

The learning-based PAD methods mainly use neural networks to implement. Neural networks can automatically learn data associations to achieve feature extraction, which has motivated researchers to use trained networks to solve PAD problems. Chugh et al. used inception-V3 to train a binary classification network and achieved 99.73% TDR @ FDR = 0.2% in the OCT patch image [38]. Liu et al. proposed a one-class

PAD method that does not require negative samples and achieves the best performance in both B-scan-wise and instance-wise cases [39]. Zhang et al. combines one-class and wavelet transform to propose a frequency-domain OCT PAD method [40]. However, these methods ignore the useful layered features that are proven effective in OCT. The network without any guidance to obtain ideal classification results requires a large number of training samples. Intuitively, considering the complexity and variety of PA samples, it is extremely challenging to increase the generalization ability of the PAD method with the premise of limited PA dataset.

To solve this challenge, we propose a novel supervised learning-based OCT PAD network method which combines effective guidance in this paper. It greatly improves the accuracy of PAD without the need for a large number of training samples. The proposed network denoted as ISAPAD, contains a dual-branch architecture. The global-feature branch accepts raw OCT images to learn the global correlations. The local-attention branch concentrates on layered structure feature which comes from the internal structure attention module (ISAM). Channel-wise connections and interactions are used in the global-feature branch and local-attention branch, so that the dual-branch architecture can promote and optimize each other. Moreover, B-scan-wise and instance-wise PAD score generation rules are employed.

The ISAM enables the proposed network to obtain layered segmentation features (including stratum corneum, viable epidermal contour, sweat glands) belonging only to Bonafide directly from noisy OCT volume data. It is worth noting that although both ISAM and work [30], [41] have segmented OCT data, the purpose of pursuit is different. Accurate segmentation is necessary for reconstructing OCT fingerprints. Thus, in the work [30], complex and diverse structures are used in pursuit of more accurate region segmentation results. In the work [41], the configuration of multi-task learning is to find the contours between different layers more precisely. In contrast, PAD does not require accurate segmentation results, so ISAM only uses the simplest structure to provide attention guidance for the approximate effective area of PAD. Therefore, single-side supervised learning is performed on ISAM to ensures that ISAPAD only obtains PAD effective regions of Bonafide during training. Although accurate segmentation results may also be effective for OCT PAD, the increased complexity may not be acceptable for PAD deployment.

In summary, a network combined with finger internal structure attention is proposed for the OCT fingerprint PAD, denoted as ISAPAD. The main contributions of this paper are summarized as follows.

1) It is the first time that an internal structure attention mechanism is proposed for OCT fingerprint PAD. This attention mechanism forces the network to focus on regions that really contribute to PAD. The proposed method greatly improves the accuracy of PAD without the need for a large number of training samples.

2) The proposed ISAM and dual-branch structure realizes direct classification from noisy data. Even in the face of complex PAs and Bonafides from different domains, the proposed architecture is still verified to be the most effective.

The remainder of this paper is organized as follows. Section II presents the proposed methods for PAD. Section III discusses the experiments and results. Section IV draws the conclusion.

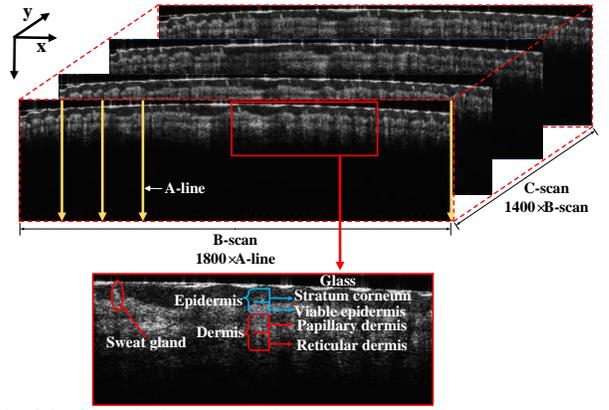

Fig. 2. OCT fingertip data presentation.

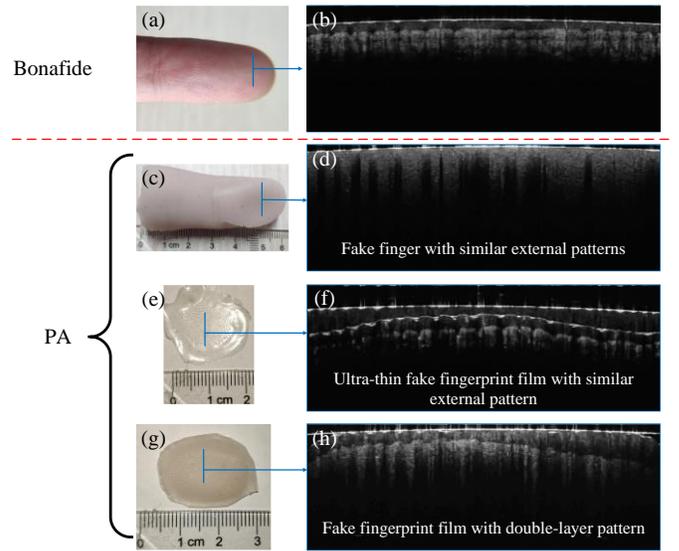

Fig. 3. (a) Bonafield. (b) B-scan corresponding to (a). (c) Fake finger with similar external pattern. (d) B-scan corresponding to (c). (e) Ultra-thin fake fingerprint film with similar external pattern. (f) B-scan corresponding to (e). (g) Fake fingerprint film with double-layer pattern. (h) B-scan corresponding to (g). Complex PAs such as (e) and (g) can be made using inexpensive and common silicone. This type of PA with an internal structure makes it difficult to distinguish from Bonafide.

## II. PAD USING ISAPAD

### A. PAD Problem Analysis

OCT imaging technology is a non-invasive optical imaging technology that can achieve high-resolution, 3D imaging of biological tissues. It is essentially based on the principle of broadband optical interferometry to image the internal structure of the finger. The light reflected from different depths of the finger skin interferes with the reference light reflected by the specular surface, and the internal structure of the finger can be analyzed from the interference signal. It performs 2D lateral scanning (B-scan) on the surface of the fingertip. From the 3D internal structure (X-Y-Z) of the fingertip, PAD features can be obtained. As shown in Fig. 2, multiple cross-sectional images are obtained to form a 3D data presentation. Meanwhile, the inner viable epidermis contour is obvious. The sweat gland is spread between the stratum corneum and the viable epidermis.

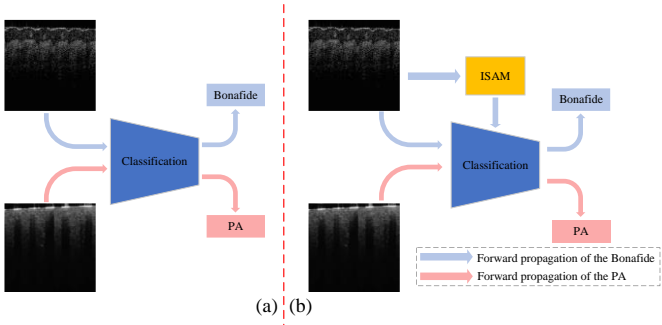

Fig. 4. The strategy that ISAM incorporates into the network.

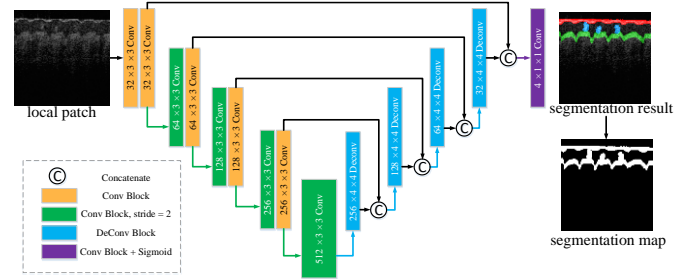

Fig. 5. The architecture of ISAM. Using the end-to-end U-Net structure, a 4-channel segmentation map is obtained. The segmentation map shows the foreground portion that is segmented and finally used for attention.

Fig. 3 illustrate some typical examples of Bonafide and PA. Compared to the real finger (Bonafide) with a clear internal layer, the internal structure of PA is diverse. Conventional OCT PAD methods focus on distinguishing between B-scan types in Fig. 3(b) and Fig. 3(d). However, PAD systems configured with such approaches are vulnerable to PAs of type Fig. 3(e) and Fig. 3(g) in field deployment. It is mainly caused by such two factors. First, PAs with type Fig. 3(e) and Fig. 3(g) contain a subset of Bonafide (Fig. 3(f)) or have a layered structure like Bonafide (Fig. 3(h)). OCT B-scan for this type of PA has a similar internal structure distribution to Bonafide. Such similar layered information may lead to false characteristics that can lead to misjudgments. Second, with the development of technology and materials, the complexity and diversity of PA samples will increase. We simply divide the PA samples into two categories: external pattern simulated and structure simulated. Building a unified model that can describe all PAs is still a difficult task.

Therefore, to extract the discriminating features of Bonafide and combine them into the network to obtain the feature space of Bonafide more accurately, we propose an ISAM to assist network learning and generalization. The strategy that ISAM incorporates into the network are shown in Fig. 4. Compared to traditional PAD classification networks (Fig. 4(a)), the forward propagation of the Bonafide adds ISAM and performs single-side supervised learning (Fig. 4(b)). The ISAM is trained to make the acquisition of internal structural features only satisfy for the Bonafide but not suit for the PA under the single-side supervise learning to obtain a compact feature space of Bonafide.

### B. ISAPAD Architecture

A novel dual-branch architecture which is denoted as ISAPAD is proposed for OCT fingerprint PAD achieving directly from noisy OCT data. It can be simply divided into 4 core structures.

*1) Internal Structure Attention Module (ISAM)*

The key to realize the proposed method is the internal structure attention module (ISAM). It is essentially a simple semantic segmentation network in the shape of a U-Net [42] that predicts individual structural layers from a local patch of noisy fingerprint OCT data. Due to the single-side supervised learning is performed, ISAM can easily return the results of layered semantic segmentation of individual fingerprint structure information. However, unlike conventional networks used for segmentation, ISAM is designed to provide potentially attentional information rather than accurate segmentation results. In the pursuit of higher precision, the architecture of the network becomes more complex, thus ISAM has only a simple structure which is shown in Fig 5. ISAM only provides guidance on the Bonafide category for classified networks, making the network more generalized to the Bonafide feature space. Using single-side supervised training to enhance the generalization of classification is effective for similar problems, such as the detection of retinopathy [43].

As shown in Fig. 5, ISAM is an end-to-end U-Net full-pixel semantic segmentation network. The targets of segmentation were the epidermis contour layer, the viable epidermal contour, sweat glands, and background in Bonafide, respectively. Each Conv Block contains two 3×3 convolution layers and one 1×1 convolution layer, using residual connection. Each deconvolution block contains one 4×4 deconvolution layer, one 3×3 convolution layer, and one 1×1 convolution layer, using residual connections. Finally, a convolutional layer of 1 layer 1×1 is used to classify the number of channels as 4, corresponding to each segmentation target one-to-one. Each convolution operation includes the batch normalization (BN) layer and ReLU activation functions. The Sigmoid function is used to map the value of the final segmentation result between 0 and 1. Channel-wise skip connections enable the shallow network features in the encoding process to be reused for segmentation detail recovery in the decoding stage. The result of the final network prediction is 4 semantic segmentation maps, representing the positions of the pixel-level foreground and background respectively. The foreground includes 3 segmentation maps, representing the positions of the stratum corneum, viable epidermis, and sweat glands respectively.

After segmentation result are estimated at ISAM, an attention strategy is performed on input image and predicted foreground segmentation map to obtain an attention image. The attention strategy is as follows:

$$ISAM(x) = S_{out}(x) \circ x * w_1 + [1 - S_{out}(x)] \circ x * w_2 \quad (1)$$

where $S_{out}(*)$ is the predicted foreground segmentation map, $1-S_{out}(*)$ represents the predicted background are. $x$ is the input image. The attention weight $w_1$ is set to 1 and the non-attention weight $w_2$ is empirically set to 0.5. ∘ is hadamard product.

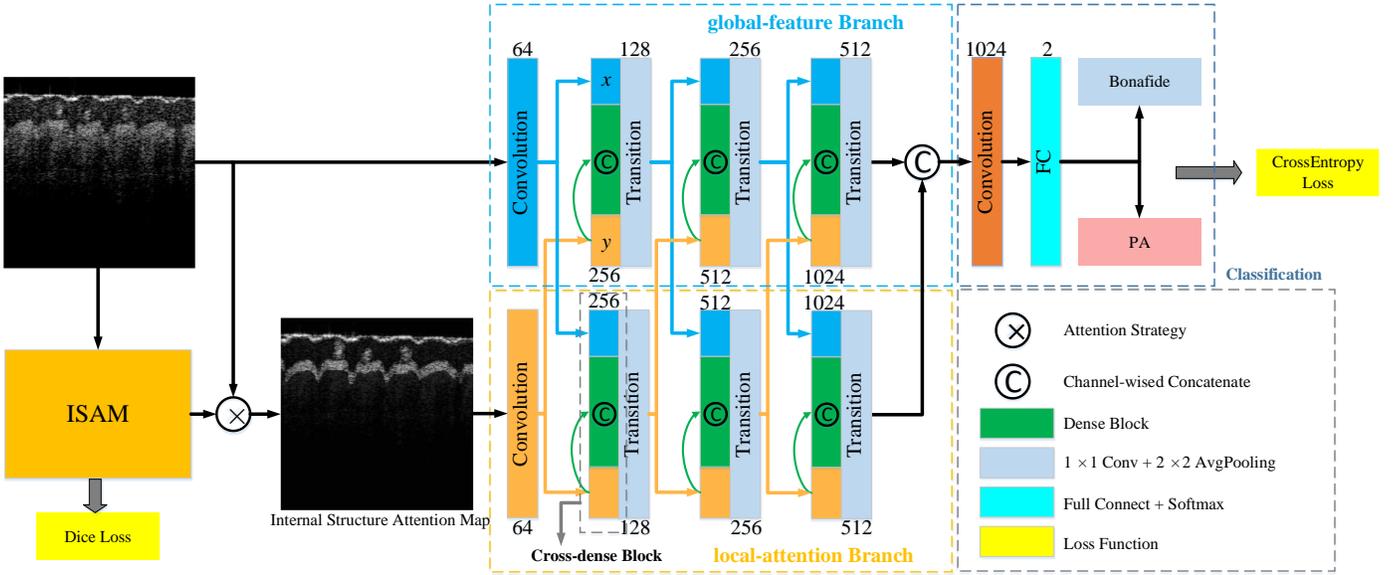

Fig. 6. Overall framework for ISAPAD. The global-feature branch and the local-attention branch both contains cross-dense block to automatically interact and transmit potential classification information.

TABLE I
THE DETAILS OF PROPOSED NETWORK

| Layers | Local-attention branch | Global-feature branch | Output size | Output channel |
|---|---|---|---|---|
| Input image | / | | $256 \times 256$ | 1 |
| ISAM | ISAM | / | $256 \times 256$ | 4 |
| Convolution | $7 \times 7$ conv, stride=2 | $7 \times 7$ conv, stride=2 | $128 \times 128$ | 64 |
| Cross-dense block1 | $x + y + \left\{\begin{array}{c}3\times 3\,\text{conv}\\1\times 1\,\text{conv}\end{array}\right\} \times 4\ for\ x^{*}$ | $x + y + \left\{\begin{array}{c}3\times 3\,\text{conv}\\1\times 1\,\text{conv}\end{array}\right\} \times 4\ for\ y$ | $128 \times 128$ | 256 |
| Transition1 | $1 \times 1$ conv | $1 \times 1$ conv | $128 \times 128$ | 128 |
| | $2 \times 2$ average pooling, stride=2 | $2 \times 2$ average pooling, stride=2 | $64 \times 64$ | 128 |
| Cross-dense block2 | $x + y + \left\{\begin{array}{c}3\times 3\,\text{conv}\\1\times 1\,\text{conv}\end{array}\right\} \times 8\ for\ x$ | $x + y + \left\{\begin{array}{c}3\times 3\,\text{conv}\\1\times 1\,\text{conv}\end{array}\right\} \times 8\ for\ y$ | $64 \times 64$ | 512 |
| Transition2 | $1 \times 1$ conv | $1 \times 1$ conv | $64 \times 64$ | 256 |
| | $2 \times 2$ average pooling, stride=2 | $2 \times 2$ average pooling, stride=2 | $32 \times 32$ | 256 |
| Cross-dense block3 | $x + y + \left\{\begin{array}{c}3\times 3\,\text{conv}\\1\times 1\,\text{conv}\end{array}\right\} \times 12\ for\ x$ | $x + y + \left\{\begin{array}{c}3\times 3\,\text{conv}\\1\times 1\,\text{conv}\end{array}\right\} \times 12\ for\ y$ | $32 \times 32$ | 1024 |
| Transition3 | $1 \times 1$ conv | $1 \times 1$ conv | $32 \times 32$ | 512 |
| | $2 \times 2$ average pooling, stride=2 | $2 \times 2$ average pooling, stride=2 | $16 \times 16$ | 512 |
| Concatenate | $x + y$ | | $16 \times 16$ | 1024 |
| Classification layer | $3 \times 3$ conv, stride=2 | | $8 \times 8$ | 1024 |
| | Global average pooling | | $1 \times 1$ | 1024 |
| | 1000D full-connected and softmax | | / | 2 |

* $x$ represents the output of the previous layer of local-attention branch. $y$ represents the output of the previous layer of global feature branch. $+$ represents the channel-wised concatenate.

*2) Dual-branch Architecture*

The basic architecture used is a DenseNet [44] that is widely used in classification tasks. Considering that the output of ISAM is configured with an attention strategy at the same scale as the input image to force neurons to extract PAD features to focus on potential internal structural regions, we divide conventional DenseNet into two branches to better combine the global region feature and the local attention area feature. As shown in Fig. 6, the proposed network architecture includes two branches, i.e. global-feature branch and local-attention branch. The global-feature branch receives the original local patch. The local-attention branch integrates with ISAM. The two branches perform convolution and transition independently. The details of the ISAPAD are shown in the Table I. The network can directly receive noisy B-scan images as input. The output is the probability of Bonafide and PA. The features of the middle layer interact only in a cross-dense block.

*3) Cross-dense Block*

To allow the two branches to automatically interact and transmit potential classification information, we reduce the number of channels generated by the traditional dense block and replace it with the previous layer feature of the other branch. Through the channel-wised concatenate, new features are generated that are identical to the original DenseNet channel. Such a structure we denoted as a cross-dense block.

| Scheme: Adaptive patch extraction |
|---|
| **m, n** are strides of local patch selected in each B-scan, are set to 256 and 64, respectively.<br>$Z_s$ is the ordinate of alternative patch center.<br>**patch**$\{I, x, z\}$ is a 256 × 256 image patch from image $I$ with $(x, z)$ center coordinates.<br>**sum()** is the sum of all pixels in an image.<br>**kernel** is a 5×5 rectangle structural element<br>**T** is a threshold set to 0.01.<br>$D(x, y, z)$ is the currently processed noisy OCT volume data. $x \in (0, X], y \in (0, Y], z \in (0, Z]$, X is the width of B-scan, Y is the total number of B-scan in a volume data, Z is the height of B-scan<br>1: Load currently processed noisy OCT volume data $D$<br>2: **for** $y$ = 1 to Y **do**<br>3:    obtain currently processed B-scan $I_s = D(:, y, :)$<br>4:    obtain $I_d$ by morphological expansion operations with **kernel** on $I_s$.<br>5:    obtain binarized image $I_b$ by performing Otsu's method on $I_d$.<br>6:    update $Z_s = \mathrm{argmax}(\sum_{x=1}^{1800} I(x,:))$ .<br>7:    **for** $z = Z_s$ to Z with stride **m do**<br>8:       **for** $x$ = 128 to X with stride **n do**<br>9:          **if sum(patch**$\{I_b, x, z\}$**)**>256 × 256 ×**T do**<br>10:             save **patch**$\{I_s, x, z\}$ for training<br>11:          **end if**<br>12:       **end for**<br>13:    **end for**<br>14: **end for** |

| Training strategy1 Independent training strategy for our network |
|---|
| 1. Pretrain ISAM only for Bonafide with *loss1*.<br>2. Fix ISAM, update ISAPAD (without ISAM) with *loss2*. |

| Training strategy2 Step-by-step alternating training strategy for our network |
|---|
| 1. Pretrain ISAM only for Bonafide with *loss1*.<br>2. Fix ISAM, update ISAPAD (without ISAM) with *loss2*.<br>3. Fix ISAPAD (without ISAM), update ISAM only for Bonafide with *loss1*.<br>4. Alternately perform steps 2 and 3. |

| Training strategy3 Joint training strategy for our network |
|---|
| 1. Pretrain ISAM only for Bonafide with *loss1*.<br>2. Fix ISAM, update ISAPAD (without ISAM) with *loss2*.<br>3. update ISAPAD with *Loss=loss1+loss2*.<br>4. Alternately perform steps 2 and 3. |

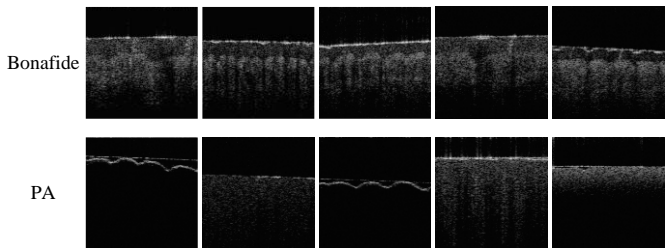

Fig. 7. The illustration of some patches by adaptive patch extraction method.

*4) Concatenate Classification*

The high-level features of the dual-branch output first perform channel-wise concatenation and feed into a classifier. The concatenate classification contains a convolution that handles concatenated high-level features, a global average pooling layer, and a fully connected layer consistent with the number of channels. For the two classification results of PAD, the final output of the multiple neuron is mapped from 0 to 1 using the Softmax function.

*C. Implementation Details of ISAPAD*
*1) Training Data Generation and Augmentation*

For each B-scan of OCT, the invalid information (mostly with a black background) occupies most of the image, directly reducing the scale of the image results in further compression of the valid information. In addition, the difference between Bonafide and PA from surface to internal structure is found throughout the OCT volume data, and the use of locally valid images is sufficient to include this potential difference. Therefore, we use an adaptive patch extraction method to generate training data from B-scan. Some patches generated for training by the adaptive patch extraction method are shown in the Fig. 7. It indicates that the patch of 256 × 256 is enough to encompass all depth information from the surface to the inside.

In the network training stage, all OCT data for training are generated using the adaptive patch extraction method.

*2) Network Training Strategy*

The internal structural differences of Bonafide and PA in the same way as the traditional feature-based PAD method could be addressed by introducing ISAM, while the potential contribution of global features and local features to classification could be intelligently learned by the introducing the dual-branch structure. Given that the entire network consists of two separate parts, a fixed ISAM may not be able to learn anything about classification. To distribute attention weights and regions more effectively, we summarize three different training strategies [45], [46], [47] and compare their applicability in experiments.

Training strategy1 is the most conventional training mode. The training of the network is divided into two separate steps. When training ISAPAD, ISAM is frozen and does not participate in training. Due to the training processes are independent of each other, ISAM cannot obtain potential classification task optimization.

Training strategy2 is a step-by-step training method. The alternating training mode allows ISAPAD to get more ISAM output at different epochs. While the output of different ISAM may make ISAPAD more generalizable, ISAM optimization is still not obtained from the *loss2* for classification task.

In fact, the expectation for ISAM is to find as many unifying features as possible that belong only to Bonafide. Therefore, the joint training strategy could encourage the network to intelligently assign attention weights on each branch based on the potential contribution of ISAM. Intuitively, ISAM and ISAPAD could be trained jointly. As shown in Training strategy3, *loss1* and *loss2* are combined in training, and progressive alternating training is performed to ensure the convergence of the network.

The dice loss function [48] is performed in ISAM training as *loss1*, which can be defined as

$$loss1 = 1 - \frac{\sum_{l}^{L}\sum_{i}^{N} p_l(i)g_l(i)}{\sum_{l}^{L}\sum_{i}^{N} p_l(i) + \sum_{l}^{L}\sum_{i}^{N} g_l(i)} \quad (2)$$

where *N* represents the all pixels of Bonafide patch, *L* represents the semantic segmentation category. $p$ and $g$ represents the predicted binary segmentation result and the label of Bonafide patch, respectively.

The CrossEntropyLoss function in PyTorch is performed in ISAPAD training as *loss2*, which can be defined as

$$loss2 = -\log \frac{\exp[y(class)]}{\sum_{j \in C} \exp[y(j)]} \quad (3)$$

where $C$ is the classification number of Bonafide and PA. $y(j)$ represents the estimated probability of $j$ class on input patch. $y(class)$ represents the estimated probability of corresponding label class on input patch.

Integrating loss1 and loss2, the objective of the network for PAD is

$$Loss = \lambda_1 loss1 + \lambda_2 loss2 \quad (4)$$

where $\lambda_1$ and $\lambda_2$ are the balanced parameters.

### D. PAD Score Generation

The advantages of using local patch for training also include the flexibility of the PAD implementation for B-scan-wise and instance-wise OCT data. We use the rule of means (which can also be thought of as the sum rule [49]) to obtain the PAD score.

*1) B-scan-wise PAD Score*

Given a set of B-scan samples, $S = \{x_1,...,x_i,...,x_N\}$, where *N* is the number of B-scan sample. Each $x_i$ can get multiple patches through the adaptive patch extraction method, $P_i = \{p_1^{x_i},...,p_j^{x_i},...,p_{N_{P_i}}^{x_i}\}$, where $N_{P_i}$ represents the total number of local patch generated by the adaptive patch extraction method. Each $p$ can get a ISAPAD output, $Z_i = \{z_1^{x_i},...,z_j^{x_i},...,z_{N_{P_i}}^{x_i}\}$. The B-scan-wise PAD Score is given by

$$Score_{PAD}^{Bscan} = \frac{1}{n}\sum_{k=1}^{n} z_k^{x_i} \quad (5)$$

Where $n$ represents that $n$ local patches are randomly sampled from $P_i$ to reduce the computational complexity. The number of random samples $n$ is empirically set to 10.

*2) Instance-wise PAD Score*

Given a set of instance samples (OCT volume data), $S = \{X_1,...,X_k,...,X_N\}$, where *N* is the number of OCT volume data. Each $X$ consists of multiple B-scans, $X_k = \{x_1^{X_k},...,x_i^{X_k},...,x_{N_{X_k}}^{X_k}\}$. To simplify the wasted time of iterating through all B-scans in an instance, we randomly sample the B-scans of each instance, $R^{X_k} = \{r_1,...,r_m,...,r_M\}$, where *M* is the number of samples for an instance. Similarly, for each $r$, multiple patches can be obtained, $P_i = \{p_1^{r_m},...,p_j^{r_m},...,p_{N_{P_i}}^{r_m}\}$. The corresponding ISAPAD output can be defined as $Z_i = \{z_1^{r_m},...,z_j^{r_m},...,z_{N_{P_i}}^{r_m}\}$. The instance-wise PAD Score is given by

$$Score_{PAD}^{instance} = \frac{1}{n \times M}\sum_{k=1}^{n \times M} z_k^{r_m} \quad (6)$$

Where the number of random samples $n$ is set to 2. In the experiment, *M* is set to 10 for simplicity. The sampling of B-scan and patch is essentially a sparse representation of OCT volume data, which is conducive to reducing complexity and quickly generating PAD scores.

## III. EXPERIMENTS AND RESULTS

In this section, comprehensive experiments are performed to verify the effectiveness and generalization of the proposed ISAPAD. The open source deep learning framework Pytorch is used to implement ISAPAD. In the training stage, Adam optimizer and adaptive learning rate at $10^{-4}$ to $10^{-5}$ are adopted. The balanced parameters, $\lambda_1$ and $\lambda_2$, are fixed empirically as 0.001 and 1, respectively.

### A. Dataset Description

Public OCT fingerprint databases ZJUT-EIFD [50] is utilized to evaluate the effectiveness of our method. It contains 3551800 B-scans from 399 different fingers of 60 people. Besides, the same OCT equipment [23] was used to collect data from the PA samples. PA data consists of 73500 B-scans from 105 different PA subjects of 21 materials. Consider a more general case, the failure of a PAD method is most likely caused by an unknown PA. The generalization ability of the model is the key to field deployment and application. Therefore, considering the limited samples, we set up an extreme training scenario to verify the generalization ability of our method. As shown in Table II, for Bonafide, just 3 people's 24 fingers (72 volume data) were used for training. Correspondingly, 24 subjects of 12 PA materials (a total of 72 volume data) were used for training to balance the positive and negative sample sizes. As previously described, PAs made with multiple materials are divided into two categories, external pattern simulated and structure simulated. For traditional AFRS, PAs with external similar surface patterns can potentially gain false access. For OCT fingerprint system, benefited to the ability of internal penetrating imaging, PAs with internal structures similar to real finger more likely spoof the system. To better test

TABLE II
THE DATA PARTITION OF PAD DATASET

| Training and validating Data | | | Data for PAD Performance |
|---|---|---|---|
| Cross-validation | Bonafide | PA | Bonafide: 3283000 B-scans from 375 fingers of 57 people<br>PA: 567000 B-scans from 81 samples of 21 PA material types |
| Training | 2×8×3×1400=67200 B-scans* | 8×2×3×1400 = 67200 B-scans** | |
| Validating | 1×8×3×1400=33600 B-scans | 4×2×3×1400=33600 B-scans | |

*The numbers in the equation that represent Bonafide are, in order, the number of people, the number of fingers collected per person, the number of acquisitions per finger, and the number of B-scans included in each acquisition.
**The numbers in the equation representing PA are, in order, the number of types of PA material, the number of per type, the number of acquisitions per sample, and the number of B-scans included in each acquisition.

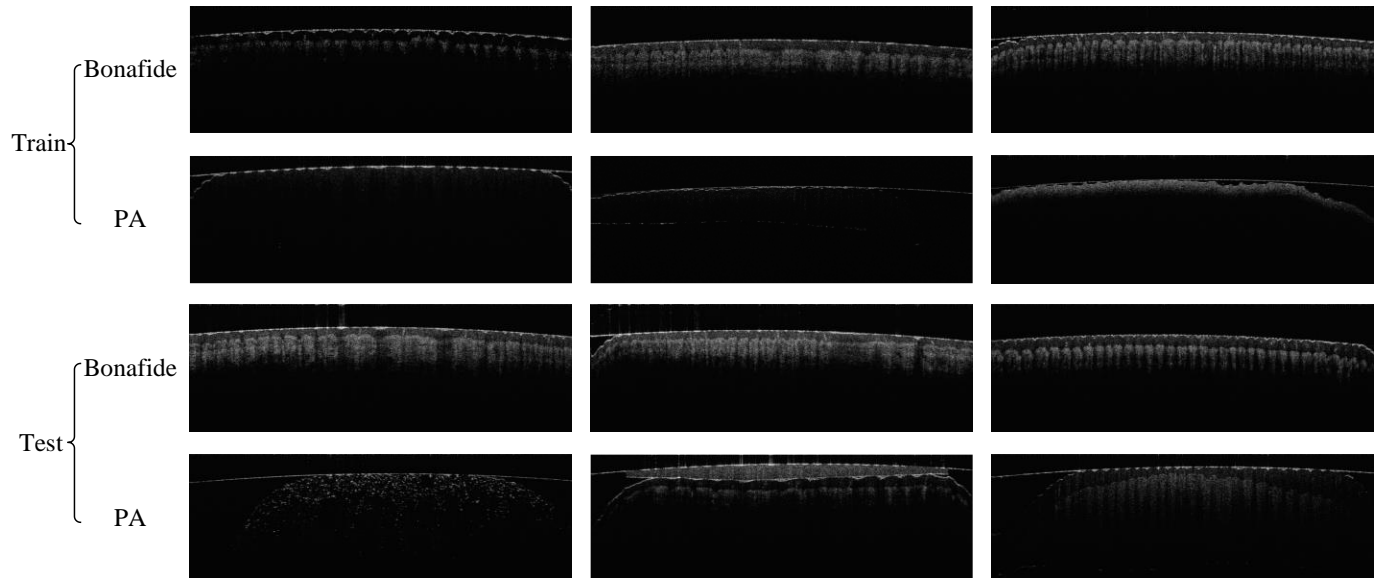

Fig. 8. The illustration of some B-scans for training and testing.

the generalization performance, the PAs used for training are all subjects with external pattern simulated, PAs with similar structures are tested for PAD performance as unknown samples. Fig. 8 shows some samples used for training and testing.

The Detection Error Tradeoff (DET) curve are reported to evaluate the performance. The DET curve plots the False Positive Rate (FPR) against False Negative Rate (FNR). For the same FPR, the lower FNR indicates the higher PAD performance. Moreover, the Equal Error Rate (EER), Half Total Error Rate (HTER) and Area Under Curve (AUC) are used as the evaluation metrics. 3-fold cross-validation is performed on a per-material basis for PA materials and different individuals. The results reported in this paper are obtained by 3-fold averaging.

*B. Assessment of PAD Performance*

To verify the performance of ISAPAD more comprehensively, a horizontal comparison of the current mainstream supervised network is performed. It including VGG16 [51], ResNet [52], InceptionV3 [53], Attention56 [54], ResNeXt [55], MobileNetV2 [56], ShuffleNetV2 [57]. The proposed adaptive patch extraction method and PAD score generation method is configured for all supervised networks to make the comparison more intuitive. In addition, feature-based OCT PAD methods are also included for comparison. The PAD performance results are shown in Fig. 9 and Table III. It should be noted that unsupervised methods that do not require PA sample training, such as OCPAD [39] or FFDPAD [40], are not compared with this method because of limited training data and data imbalance problems.

From the result, it can be observed that comparing to other supervised-based models, our method achieves optimal performance in both B-scan-wise and instance-wise PAD results. On the B-scan-wise scale, ISAPAD was 28.6% lower than the second-rank value on EER, 23.9% lower on HTER, and 0.7% higher on the AUC. On the instance-scan-wise scale, ISAPAD was 43.8% lower than the second-rank value on EER, 43.5% lower on HTER, and 0.7% higher on the AUC. The excellent performance of our method can be summarized as: ISAPAD makes feature extraction simpler and more efficient. Specifically, all other supervision-based models lack the ability to proactively focus on internal structural differences. In the case of limited training sets (Bonafide: 2 individuals, PA: 8 materials and only with external pattern simulated), only features biased towards known PAs can be extracted. In fact, the diversity of PA makes it unrealistic to fully generalize the PA feature space, and overfitting the existing PA may affect the classification results of Bonafide. In contrast, benefit from the traditional feature-based PAD approach, the search for a unified Bonafide feature space clearly contains approximate internal structures. Therefore, ISAPAD can provide the guarantees for the generalization performance of Bonafide.

In addition, as mentioned before, feature-based PAD methods, such as autocorrelation-based method [37] and peak-based method [32], aim to find generalized feature spaces for both PA and Bonafide. However, the complexity of PAs makes it difficult to summarize the characteristics of PAs. Insufficient discriminative capability of features will lead to ineffective detection of unknown PA.

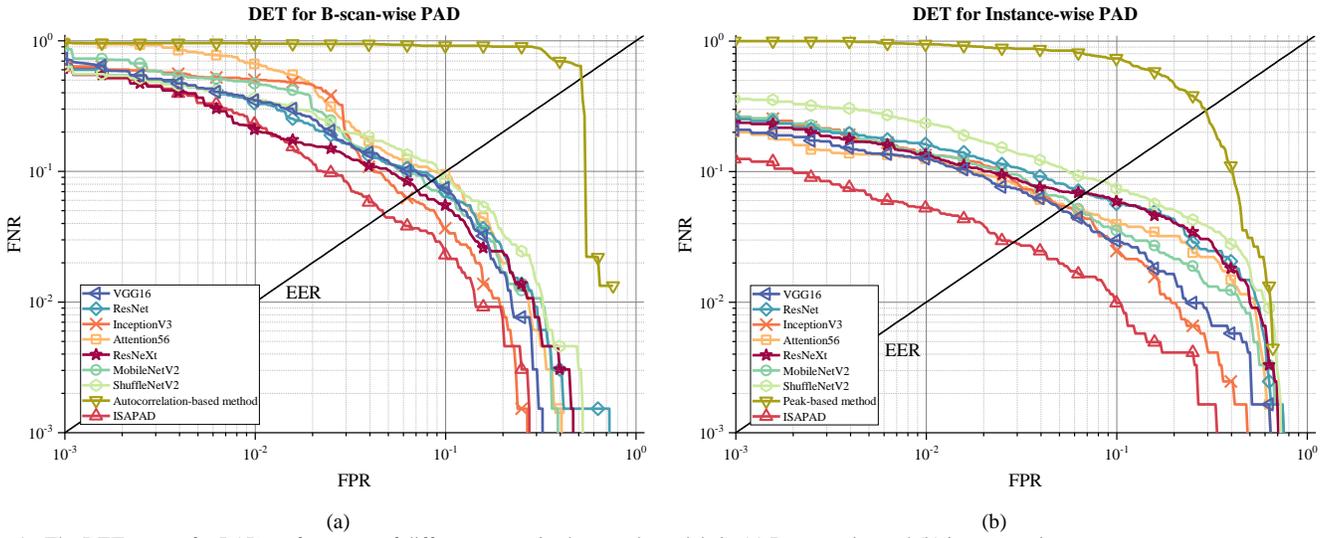

Fig. 9. The DET curves for PAD performance of different supervised network models in (a) B-scan-wise and (b) instance-wise.

TABLE III
THE PERFORMANCE OF DIFFERENT PAD METHODS (%)

| PAD Methods | B-scan-wise | | | Instance-wise | | |
|---|---|---|---|---|---|---|
| | EER | HTER | AUC | EER | HTER | AUC |
| VGG16 [51] | 8.436 | 12.670 | 97.794 | 5.126 | 13.743 | 98.876 |
| ResNet [52] | 8.747 | 11.554 | 97.631 | 6.831 | 13.609 | 97.682 |
| InceptionV3 [38] [53] | 6.269 | 8.264 | 97.844 | 5.395 | 11.539 | 98.950 |
| Attention56 [54] | 9.633 | 16.817 | 96.729 | 5.432 | 13.959 | 98.286 |
| ResNeXt [55] | 7.117 | 9.992 | 98.156 | 6.849 | 12.717 | 97.780 |
| MobileNetV2 [56] | 7.844 | 10.697 | 97.473 | 5.776 | 11.202 | 98.438 |
| ShuffleNetV2 [57] | 9.497 | 11.355 | 97.045 | 8.417 | 13.362 | 96.847 |
| Autocorrelation-based method [37] | 49.197 | 44.083 | 56.776 | / | / | / |
| Peak-based method [32] | / | / | / | 29.359 | 31.599 | 79.247 |
| ISAPAD | **4.475** | **6.288** | **98.876** | **2.881** | **6.328** | **99.621** |

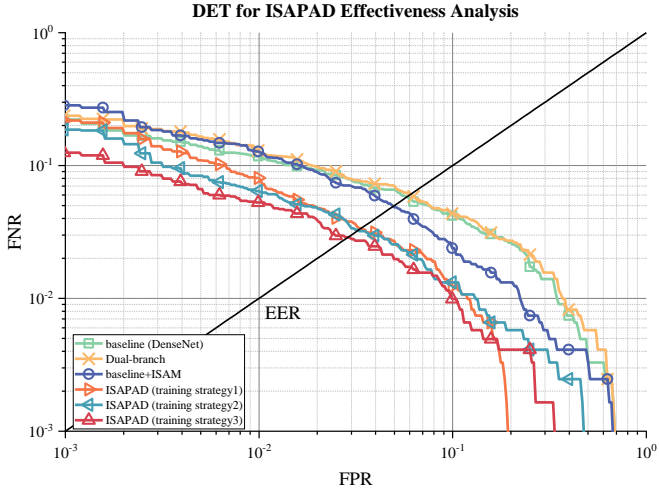

Fig. 10. The DET curve for effectiveness analysis. It includes ablation experiments and comparisons of different training strategies.

TABLE IV
EFFECTIVENESS ANALYSIS RESULTS OF ISAPAD (%)

| | Models | EER | HTER | AUC |
|---|---|---|---|---|
| Ablation Experiments | Baseline (DenseNet) | 5.844 | 15.308 | 98.532 |
| | Dual-branch | 6.027 | 11.499 | 98.383 |
| | Baseline + ISAM | 4.930 | 8.281 | 98.965 |
| Training Strategies | Training Strategy1 | 3.370 | 6.804 | 99.538 |
| | Training Strategy2 | 3.251 | 8.347 | 99.468 |
| | Training Strategy3 | **2.881** | **6.328** | **99.621** |

C. Effectiveness of ISAPAD

1) Ablation Experiments

To verify the validity of the proposed structure, ablation experiment is performed. The DenseNet [44] is evaluated and labelled as 'Baseline'. After removing ISAM, ISAPAD is evaluated and marked as 'Dual-branch'. The dual-branch receives two same patches, instead of one branch is the output of ISAM. DenseNet is evaluated in combination with ISAM and labeled as 'baseline + ISAM'. The instance-wise results are shown in Fig 10 and Table IV.

From the results we can get the following observations: First, Dual-branch achieves the worst performance. In the absence of ISAM, cross-dense blocks in dual-branch essentially only increase computational complexity and do not optimize the features extracted by the network. Second, ISAM combined with baseline can achieve better PAD performance than a single baseline, which fully demonstrates the effectiveness of ISAM. Third, ISAPAD achieve optimal performance, which presents the benefit of ISAM and dual-branch architecture. ISAM makes it easier for networks to learn the consistency of internal structures. The dual-branch structure enables the network to accept raw and structure-attention image input. It indicates that the two branches automatically interact and pass global features and local features to optimize PAD results.

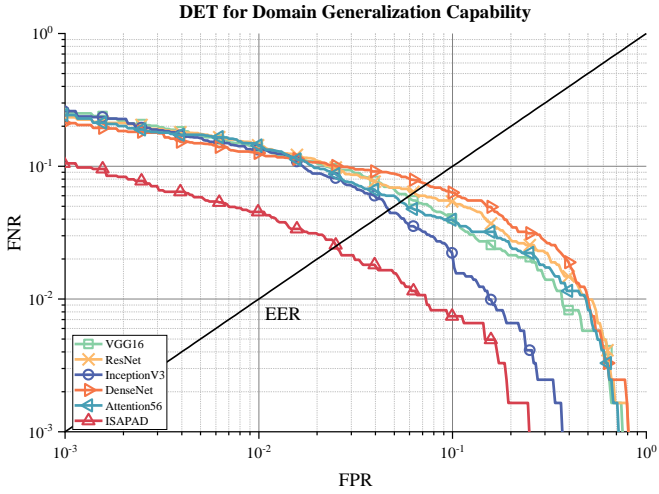

Fig. 11. (a) The DET curve for domain generalization experiment.

TABLE V
DOMAIN GENERALIZATION RESULTS (%)

| Models | EER | HTER | AUC |
| --- | --- | --- | --- |
| Vgg16 [51] | 5.852 | 13.900 | 98.440 |
| ResNet [52] | 6.329 | 13.561 | 98.054 |
| InceptionV3 [38] [53] | 4.680 | 11.533 | 99.166 |
| DenseNet [44] | 7.341 | 16.481 | 97.771 |
| Attention56 [45] | 5.395 | 14.043 | 98.383 |
| ISAPAD | **2.509** | **6.298** | **99.719** |

*2) Training Strategy Comparison*

The training strategy comparison result are shown in Fig. 10 and Table VI. As discussed in section 2.3, progressive alternating training method in Training Strategy3 allows ISAM to be optimized for classification to achieve best PAD performance. Thus, in this study, the ISAPAD with Training Strategy3 is used as the final version. The PAD performance of Training Strategy1 and Training Strategy2 are similar and exceeded that of other methods compared before, which fully demonstrated the effectiveness of ISAPAD.

*D. Domain Generalization Capability*

Due to the large distribution discrepancies of data collected by different devices, it is a challenge to the generalization ability and robustness of PAD method in the field deployment. Considering that there is no public OCT fingerprint PA database, to verify the robustness of our method, we only tested the public SZU OCT fingerprint database [58] using the model trained on our dataset. It should be noted that comparing our database, SZU OCT fingerprint dataset was established by different light source. As a result, there are large differences in image distribution between different datasets. The generalization result of instance-wise is shown in Fig. 11 and Table V.

It can be observed that ISAPAD has a significant improvement over other methods (especially the HTER metric). At the same time, the performance of ISAPAD is further improved after adding the SZU OCT database compared to the PAD performance that only uses our dataset. To more clearly indicate the generalizability of ISAPAD, we plot the distribution curve of the PAD score in both our and SZU database, the result is shown in the Fig. 12. It can be obvious that SZU Bonafide's PAD distribution is completely within our Bonafide's PAD distribution, which means that ISAPAD can accurately identify Bonafide from unknown domains. In general, the excellent performance shows the generalization capability of ISAPAD between different domains witch guarantee the effectiveness of ISAPAD in field deployment.

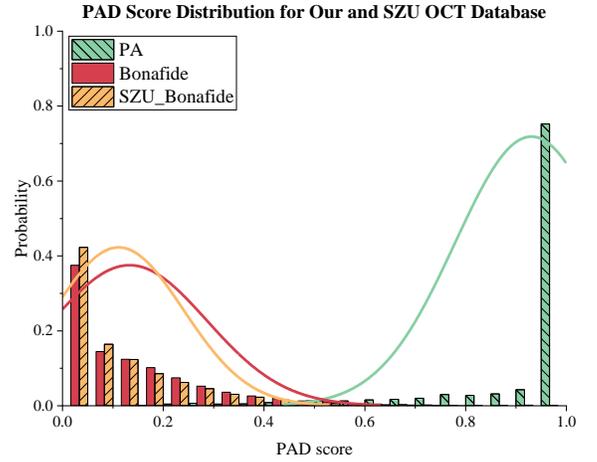

Fig. 12. Instance-wise PAD score distribution of our and SZU OCT database [58] using ISAPAD.

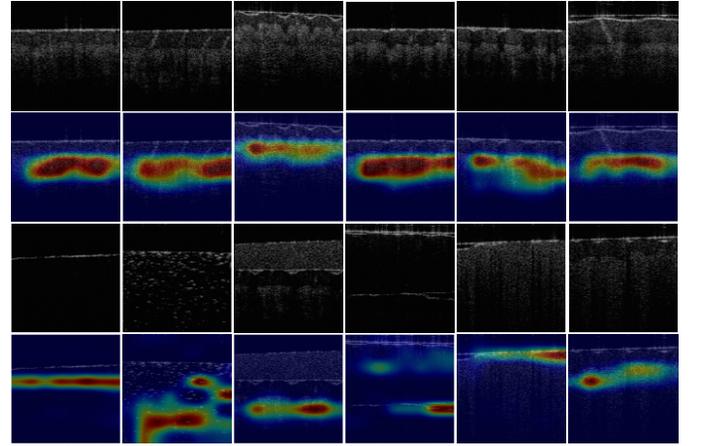

Fig. 13. Grad-CAM [59] visualizations of the ISAPAD. The first two rows show the Bonafide patch and the corresponding heat maps. The last two rows show the PA and the corresponding heat maps.

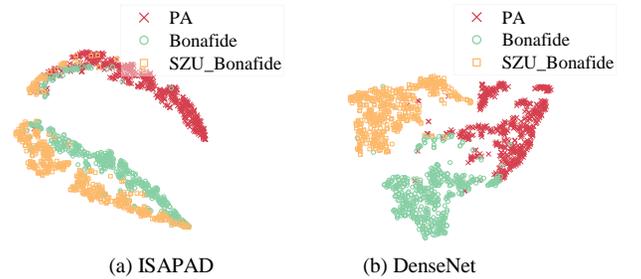

(a) ISAPAD     (b) DenseNet

Fig. 14. The t-SNE [60] dimension reduction of the PAD features by the (a) ISAPAD and (b) DenseNet.

*E. Interpretability Analysis*

As shown in Fig. 13, the Grad-CAM [59] was adopted to provide the class activation map (CAM) visualizations of our method. The results showed that ISAPAD decisions were almost exclusively derived from internal structural regions rather than superficial epidermal regions, which is more likely to generalize and convergence well to OCT fingerprint PAD task. Specifically, for PAs, our method can automatically notice differences in internal structure, resulting in higher accuracy PAD.

Furthermore, the network output features of PA and Bonafide are dimensionally reduced using t-SNE. The features presented are from ISAPAD and DenseNet respectively. 500 samples were randomly selected for each PA and Bonafide category from our and SZU OCT databases. As shown in Fig. 14, Bonafide with two different feature spaces can obtain closer distances using ISAPAD. Benefit from the single-side trained ISAM, the network is guided to search for the internal structure of Bonafide first, so that Bonafide can obtain similar PAD feature distributions in different domains.

## IV. CONCLUSION

This paper presents a novel supervised learning-based PAD method, denoted as ISAPAD, which applies prior knowledge to guide network training and enhance the generalization ability. A simple yet effective ISAM enables the proposed ISAPAD to obtain layered segmentation features belonging only to Bonafide thus greatly guide the network to focus on PAD benefited regions. The proposed ISAPAD further utilizes adaptive patch extraction method and joint training strategy to facilitate the feature extraction. For different OCT PAD scenarios of B-scan-wise and instance-wise, ISAPAD uses effective PAD strategies to ensure performance.

ISAPAD achieves optimal performance in comparison with existing supervised-based networks and existing OCT PAD methods. Ablation experiments and training strategy comparison are performed to verify the effectiveness of proposed ISAM and dual-branch architecture. Based on the domain generalization capability in our and SZU OCT fingerprint database, the effectiveness and robustness of ISAPAD can be guaranteed in field deployment. The visualization results also show that ISAPAD pays more attention to the differences in the internal structure and can make the Bonafide features from different domains more compact. In brief, the strong generalization ability and excellent PAD performance of the proposed ISAPAD are of great significance for field-deployed applications of OCT PAD.

Although the domain generalization ability of ISAPAD is validated in existing datasets, the lack of PA samples from different domains severely limits comprehensive evaluation. Therefore, further work is required to 1) establish a more comprehensive OCT PAD database to verify the robustness of the method in cross-device database evaluation, 2) while ensuring generalization performance, consider reducing the training data deployed in the field for faster configuration and implementation.